\documentclass[letterpaper]{article} % DO NOT CHANGE THIS
\usepackage{aaai23}  % DO NOT CHANGE THIS
\usepackage{times}  % DO NOT CHANGE THIS
\usepackage{helvet}  % DO NOT CHANGE THIS
\usepackage{courier}  % DO NOT CHANGE THIS
\usepackage[hyphens]{url}  % DO NOT CHANGE THIS
\usepackage{graphicx} % DO NOT CHANGE THIS
\usepackage{natbib}  % DO NOT CHANGE THIS AND DO NOT ADD ANY OPTIONS TO IT
\usepackage{caption} % DO NOT CHANGE THIS AND DO NOT ADD ANY OPTIONS TO IT
\usepackage{algorithm}
\usepackage[noend]{algpseudocode}

\usepackage{newfloat}
\usepackage{listings}
\DeclareCaptionStyle{ruled}{labelfont=normalfont,labelsep=colon,strut=off} % DO NOT CHANGE THIS
\floatstyle{ruled}
\newfloat{listing}{tb}{lst}{}
\floatname{listing}{Listing}
\usepackage{bbding}
\usepackage{pifont}
\usepackage{subfigure}

\title{SARAS-Net: Scale And Relation Aware Siamese Network for Change Detection}
\title{SARAS-Net: Scale and Relation Aware Siamese Network for Change Detection}
\author {
    Chao-Peng Chen\textsuperscript{\rm 1},
    Jun-Wei Hsieh\textsuperscript{\rm 1}*, 
    Ping-Yang Chen\textsuperscript{\rm 2}, 
    Yi-Kuan Hsieh\textsuperscript{\rm 1}, 
    Bor-Shiun Wang\textsuperscript{\rm 2}
}
\affiliations {
    \textsuperscript{\rm 1}College of Artificial Intelligence and Green Energy, National Yang Ming Chiao Tung University, Taiwan.\\
    \textsuperscript{\rm 2}Department of Computer Science, National Yang Ming Chiao Tung University Taiwan.\\
    f64051041@gmail.com, [jwhsieh, pingyang.cs08, khjhsnaughty.ai10, eddiewang.cs10]@nycu.edu.tw
}
\begin{document}

\maketitle

\begin{abstract}
Change detection (CD) aims to find the difference between two images at different times and outputs a change map to represent whether the region has changed or not. To achieve a better result in generating the change map, many State-of-The-Art (SoTA) methods design a deep learning model that has a powerful discriminative ability. However, these methods still get lower performance because they ignore spatial information and scaling changes between objects, giving rise to blurry or wrong boundaries. In addition to these, they also neglect the interactive information of two different images. To alleviate these problems, we propose our network, the Scale and Relation-Aware Siamese Network (SARAS-Net) to deal with this issue. In this paper, three modules are proposed that include relation-aware, scale-aware, and cross-transformer to tackle the problem of scene change detection more effectively.  To verify our model, we tested three public datasets, including LEVIR-CD, WHU-CD, and DSFIN, and obtained SoTA accuracy. Our code is available at https://github.com/f64051041/SARAS-Net.
\end{abstract}

\section{Introduction}
\label{intro}
\begin{figure}
\centering
\includegraphics[width=0.45\textwidth]{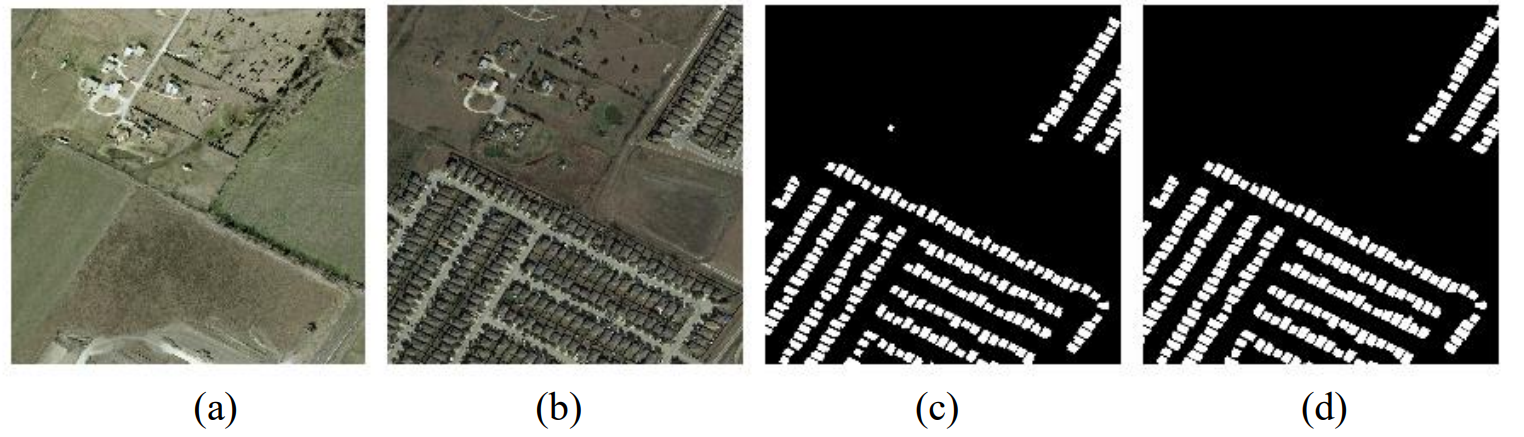}
\caption{Result of our model for change detection in the LEVIR-CD dataset. (a) and (b) input remote sensing images, (c) ground truth, and (d) prediction result of our model.}
\label{introduction}
\end{figure}

Change detection is a critical and challenging research topic in computer vision and remote sensing. This issue aims to find the difference between two images at different times and output a change map to represent whether the region has changed or not, as shown in Figure~\ref{introduction}. 
The change detection task has been widely used in many applications, such as urban expansion~\cite{lu2011detection}, damage assessment~\cite{Building_Damage}, and land cover monitoring~\cite{HULLEY2014755}. To generate a change map, most traditional methods focus on detecting the changed pixels and classifying them. However, these results often come with low accuracy because of some noise, including different light intensity and surface colors. Hence, designing a good network with powerful discrimination to solve these problems is crucial.

\begin{figure*}[t]
\centering
\includegraphics[width=1\textwidth]{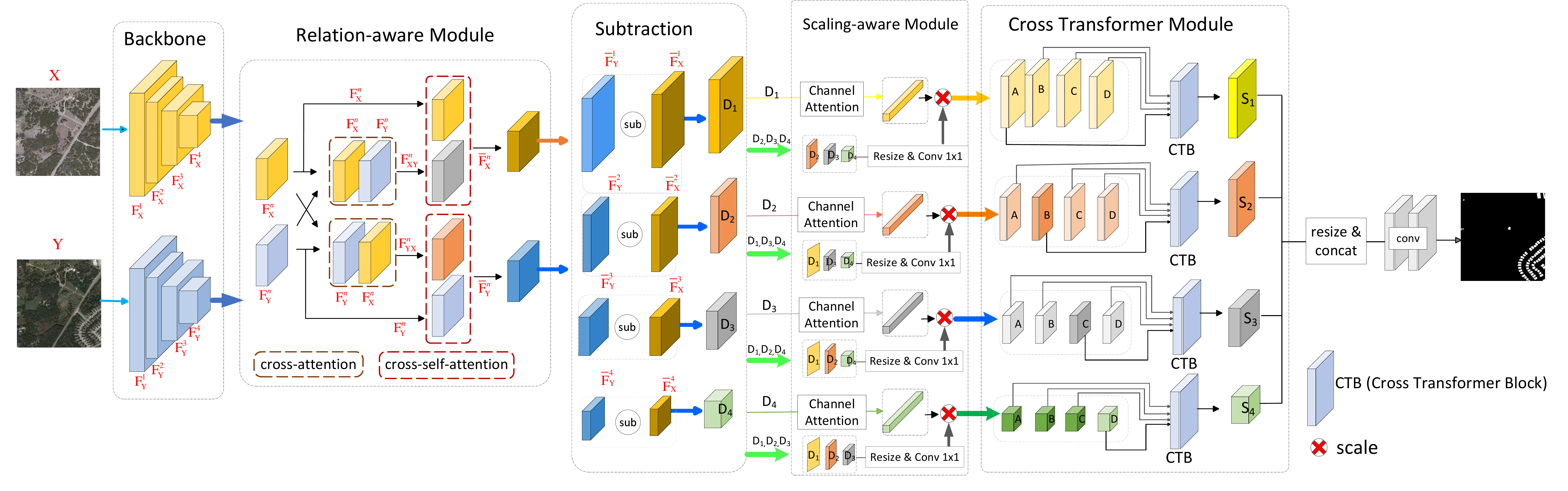}
\caption{Overview of the proposed scale- and relation-aware siamese network.}
\label{model}
\end{figure*}

With the development of deep learning, most existing methods have been proposed with powerful CNN models to tackle change detection. They have better performance than traditional methods because their outstanding discriminative ability can extract more useful features from images. However, these methods still face some problems when analyzing the change region. For example, FCN~\cite{aaa123} uses the U-net model to detect the region that constructs new buildings. Although it can roughly indicate the position of newly built constructions, it gets low performance because it ignores spatial information and different scale changes between objects. Although SNUNet~\cite{9355573} focuses on processing multi-scale features to tackle the scaling changes of objects through an ECAM (Ensemble Channel Attention Module).  However, this ECAM considers only channel attention and ignores the spatial relations between pixels to generate the change maps, so many unexpected regions with seasonal changes in vegetation are also detected.  To punish attention to unchanged feature pairs and increase attention to changed feature pairs, some methods~\cite{9311793, ZHANG2020183, 9254128} have used attention mechanisms, such as channel attention and spatial attention, to improve the detection result. However, these networks emphasize each pixel's channel importance to make the extractor more effective; it still neglects the cross-relation between features that are generated by two remote sensing images. In contrast to these networks, BIT~\cite{c2022} uses the transformer~\cite{NIPS2017_3f5ee243} to encode high-level concepts of the change of interest by a set of semantic tokens and then fuses them with the original deep features to generate the expected binary change map. Though it applies attention mechanisms and considers the relationship between two features, it does not consider performing some convolution operations to fine-tune change maps after feature subtraction. 

From the above discussions, we summarize the problems they encountered, including multi-scale objects, the relation between two images, and focusing on important channels. In addition, we find that there is another factor that influences their effectiveness. That is, all of them perform convolution before or after the distance of features. The first type of methods, for instance, FCN and SNUNet, initially concatenates two input images and then uses some convolution operations on the concatenated map to output the change result. The second type of methods, like DASNet~\cite{DASNet2021} and BIT, initially performs some convolution operations on input images and subsequently subtracts their feature maps through a few convolution layers to generate the change map. However, in our experiments, we find that performing all operations before and after feature subtraction can obtain more information and result in a better result. 

This paper proposes a new network with some mechanisms to solve the disadvantages of the above methods, as shown in Figure~\ref{model}. First, our network performs both operations before and after feature subtraction, respectively, using the relation-aware module before subtraction and using the scale-aware and cross-transformer modules after subtraction. The goal of the relation-aware module is to enhance the interactive relationships between feature maps extracted from two input images to improve the discrimination ability of features for change detection.  Then after feature subtraction, the scale-aware attention module computes cross-scaling attention on subtraction maps to deal with the problem of scene change caused by objects with multiple sizes.  Finally, the cross-transformer module, which fuses the multi-level features, aims to pay more attention to spatial information and separate foreground and background easily, thus reducing false alarms.

 To solve the change detection problem and improve features' discrimination abilities, our model contributions in this paper are as follows:
\begin{itemize}
\item We propose a siamese network that performs both operations before and after feature subtraction on two input images to detect the change region and obtain state-of-the-art performance on the remote sensing datasets.

\item We propose the relation-aware module to make the features, which are extracted before subtraction, 
have more information exchanges to improve their discrimination ability for change detection. 
\item We propose the scale-aware module, which makes the features focus on more important channels by computing cross-scaling attention on subtraction maps, to more effectively detect changes caused by different scaled objects.
\item We propose the cross-transformer module to easily separate changed pixels from unchanged ones by a self-attention mechanism. 
\end{itemize}

\section{Related Work}
\subsection{Change detection}
In the literature, many frameworks have been proposed for change detection. According to the processed units, they can be further divided into two classes, respectively, pixel-based~\cite{sdddsad00, 5196726, WU2017241} and object-based methods~\cite{GILYEPES201677, 8335352}. The first one generates a change map by comparing two coregistered images captured at different time pixel by pixel and then using a threholding or clustering method to determine the locations of changed regions. However, many false detections will be produced due to many irrelevant
changes such as lighting, weather, atmospheric, changes in road conditions, or seasonal changes in vegetation. To alleviate the above problems, some image preprocessing tasks should be performed, such as radiation correction, geometric correction, brightness normalization, and so on. Object-based methods~\cite{GILYEPES201677, 8335352} typically use structural and textural features to segment raw images into objects and then obtain their change maps. Although object-based methods consider spatial information to improve change detection performance, they are sensitive to registration errors, objects' shadows, lighting conditions, and the performance of their adopted segmentation algorithms.

Using a convolutional neural network (CNN) to extract deep
features for change detection has become more
popular in recent years and performs much better than hand-crafted features.  Basically, CNN-based methods adopt an encoder-decoder-like (or U-Net) architecture to generate the desired change map, where the encoder converts the image pairs to various feature pyramids from which the decoder generates the final change map. For example, FCN~\cite{aaa123} uses two SAR images to detect regions that include new buildings. Furthermore, an improved UNet++ architecture~\cite{rs11111382} was proposed to obtain the final binary change map by concatenating co-registered image pairs as inputs.  In addition to UNet-based methods, more Siamese architectures with various attention mechanisms were proposed for change detection.  For example,  FC-Siam-diff~\cite{aa8451652} uses a symmetric network to extract two temporal features and subtract them to obtain the change map. The difference map is the most intuitive feature to reveal the changes in bitemporal images, although the existence of spectral and position errors will produce many false alarms. Thus, more frameworks are proposed for change detection based on the difference in images.  For example, in~\cite{DASNet2021}, a dual-attentive fully convolutional Siamese Networks (DASNet) was proposed to obtain more discriminant features by focusing both channel and spatial attention together for change detection.  In~\cite{ZHANG2020183, 9254128}, attention maps were calculated not only from raw image pairs but also from their difference maps to assign changed pixels with higher importance but unchanged pixels with lower importance.  BIT~\cite{c2022} proposed an effective transformer-based change detection architecture and paid more attention to the changed regions.
\vspace{-0.2 em}
\subsection{Transformers}
\vspace{-0.1 em}

Transformer~\cite{NIPS2017_3f5ee243} is a new attention-based method for machine translation and has achieved promising performance in computer vision \cite{Rezatofighi_2019_CVPR,NEURIPS2020_9d684c58}. For example, with the Vision Transformer (ViT)~\cite{11929} as the backbone, more informative features can be extracted than using spatial convolution layers networks, such as ResNet~\cite{NEURIPS2019_3416a75f}. ViT outperforms and achieves better accuracy than CNN-based methods~\cite{NEURIPS2019_3416a75f, liu2016ssd, bochkovskiy2020yolov4} in several vision tasks including object detection~\cite{Rezatofighi_2019_CVPR} and image segmentation~\cite{NEURIPS2020_9d684c58}.  It splits the original image into non-overlapping medium-sized patches and computes their self-attentions to get more discriminant features. Although it performs well, it is very time-consuming. To alleviate this inefficiency, Swin transformer~\cite{Liu_2021_ICCV} uses a smaller window size and patch interaction mechanism to achieve better speed-accuracy trade-off in image classification.
This paper will employ this self-attention mechanism to strengthen feature maps not only at the same
scales but also cross scales to well detect areas of changed pixels with various sizes.

\section{Methodology}
\subsection{Overview}

Most SoTA methods~\cite{Chen2022Transformers,chen2020spatial,DASNet2021} used the attention module to enhance features  before image subtraction from image pairs or fewer methods~\cite{Bai2022Edge} enhanced the difference map after subtraction.  More importantly, the above methods calculated attention from features layer by layer only at the same scales.  Many miss predictions to small change areas and false alarms to large irrelevant changes will be produced. Two key ideas are proposed in this paper to alleviate the above scaling problems.  The first one is to calculate attention for enhancing features from image pairs not only before subtraction but also from the difference map after subtraction. The second one is to calculate attention from deep features layer by layers not only at the same scales but also cross scales to well detect change areas even with various sizes. Our model is shown in Figure~\ref{model} and its details are shown in Algorithm 1. To compare two temporal high-resolution remote sensing images, we design a Siamese network model to extract their features.  Firstly, a relation-aware model is proposed and applied to image pairs before subtraction to fuse feature maps and enhance their discriminant capabilities for change detection. After subtraction,  we use the scale-aware module tocalculate channel attentions on feature maps to not only at the same scale, but also other scales to deal with the scale-aware problem in change detection.  After channel weighting by the scale-aware module,  we use the cross-transformer to further cross-fuse features from different layers to captures more spatial and semantic information for detecting regions with change caused by objects with various sizes. Our model obtains SoTA performance on three public remote sensing datasets, including LEVIR-CD, WHU-CD, and DSFIN.

\subsection{Relation-aware module}
Let $X$ and $Y$ be the two input images. 
To compare $X$ and $Y$, a backbone such as ResNet~\cite{He_2016_CVPR} is first adopted to extract two feature pyramids $F_X$ and $F_Y$, respectively. Let $F_X^n$ and $F_Y^n$ denote the feature maps of $F_X$ and $F_Y$ at the $n$th layer.  To better detect changed pixels, $F_X^n$ and $F_Y^n$ will be improved
by two mechanisms, respectively, cross-attention and cross-
self-attention. Detailed operations, shown in Figure 3,
are composed of sequentially connected encoder layers. The
input features $F_i$ and $F_j$ initially produce queries $Q_i$ and $Q_j$, keys $K_i$ and $K_j$, and
value $V_i$ and $V_j$, then they are passed to the attention layer.  After generating the attention weight by the dot product between the query $Q_i$ and the key vector $K_j$, the attention information is retrieved by the product of the value vector $V_j$ and the attention weight. The attention layer is denoted as:
\begin{equation}
A(Q_i,K_j,V_j) = softmax(Q_i K_j^{T})V_j.
\label{Attention}
\end{equation}
When the attention vector is obtained, we concatenate it and the input feature $F_i$ to get a new feature $F_{i,j}$ as follows. 

\begin{equation}\label{eq:ValueAttentions} 
{F}_{i,j} ={F}_{i}+A(Q_i,K_i,V_i)+A(Q_i,K_j,V_j).
%\vspace{-0.3cm}
\end{equation}

Similarly, we can also obtain $F_{j,i}$. In the end, the output vector is computed by a $3\times3$ convolution and normalization. For the cross-attention module, $F_{i}$ and $F_{j}$ have the same size. As to the cross-self-attention module, $F_{i}$ is generated by the cross-attention module, while $F_{j}$ is from the original image. Both two modules can strengthen the features with more discriminant capabilities for change detection.

\subsection{Scale-aware module }

  After the relation-aware module, the enhanced feature maps \{${\bar{F}}_X^n$\} and \{${\bar{F}}_Y^n$\} can be obtained from $F_X^n$ and $F_Y^n$.
  By subtracting ${\bar{F}}_X^n$ with ${\bar{F}}_Y^n$, the subtraction result $D_n$ can be obtained, $i.e.$ $D_n = abs(\bar{F}_X^n-\bar{F}_Y^n$).  Given the set of difference maps \{$D^i$\}, this section will propose a scale-aware module to enhance their discriminant capabilities for change detection. Different from other attention methods which only consider attention on feature maps at the same scales, this module calculates attention on feature maps not only at the same scale but also on other scales to deal with the scale-aware problem in change detection. 
  
  First, a global average pooling is applied to $D_n$ to form a $1\times C$ vector. It is then followed by a $1\times 1$ convolution and activated a Sigmoid function to form a $1 \times C $ attention vector $U_n$. For all $\{D_n\}$, they will be resized to have the same size as $D_n$ with a bilinear interpolation operation which is followed by a $1\times 1$ convolution. For each resized $D_m$, their channels are then weighted by $U_n$ to form a new feature map $D_m^n$. For the $n$th layer, all $\{D_m^n\}$ are then sent to the cross-transformer to generate a scale-aware feature map. 
 
\begin{figure}[t]
\centering
\includegraphics[width=0.5\textwidth]{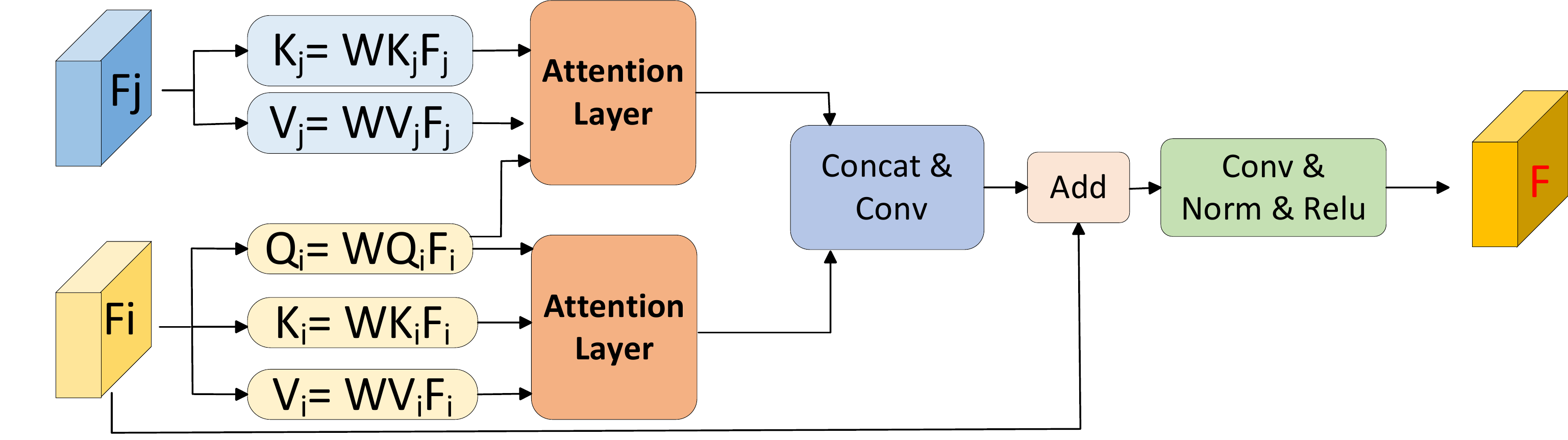}
\caption{  Detailed operations of the cross-self-attention module and cross-attention module.}
\label{relation_aware}
\end{figure}

\begin{table*}[t]
\caption{ Compare our model with other methods on LEVIR-CD, WHU-CD, and DSIFN-CD dataset}
\scalebox{0.85}{\begin{tabular}{l|ccccc|lllll|lllll}
\hline
               & \multicolumn{5}{c|}{\textbf{LEVIR-CD}}                                                      & \multicolumn{5}{c|}{ \textbf{WHU-CD}}                                                                                                      & \multicolumn{5}{c}{\textbf{DSIFN-CD}}                                                                                                    \\
Model          & Pre.           & Rec.           & F1             & IoU            & OA             & \multicolumn{1}{c}{Pre.} & \multicolumn{1}{c}{Rec.} & \multicolumn{1}{c}{F1} & \multicolumn{1}{c}{IoU} & \multicolumn{1}{c|}{OA} & \multicolumn{1}{c}{Pre.} & \multicolumn{1}{c}{Rec.} & \multicolumn{1}{c}{F1} & \multicolumn{1}{c}{IoU} & \multicolumn{1}{c}{OA} \\ \hline
FC-EF         & 84.82          & 77.55          & 81.02          & 68.11          & 97.99          & 77.24 & 68.88          &72.82         & 57.26          & 97.82                   & 61.80 & 57.75          & 59.71         & 42.56          & 86.77                  \\
FC-Siam-Di     & 86.73         & 77.52          & 81.87          & 69.31          & 98.11         & 71.61 & 73.40          & 72.49         & 56.86          & 97.64                   & \textbf{68.44} & 58.27          & 62.95         & 45.93          & 88.35                    \\
FC-Siam-Conc  & 79.85         & 83.00          & 81.39         & 68.62         & 97.90         & 76.94 & 69.74          & 73.17         & 57.69          & 97.83                  & 59.08  & 62.80          & 60.88         & 43.76          & 86.30                  \\
STANet         & 89.47          & 83.31         & 86.28         & 75.88          & 98.54     & 90.62                    & 86.26                    & 88.38                  & 79.19 & 99.04                   & 51.48 & 36.40          & 42.65         & 27.11          & 83.38                \\
IFNet          & 83.12 & 79.58          & 81.31          & 68.51          & 97.98          & 75.92           & 71.53                  & 73.66               & 58.31                   & 97.83       & 63.75                & 55.36                    & 59.26    & 42.11              & 87.08                  \\
SNUNet         & 89.43          & 87.72          & 88.57         & 79.48         & 98.75         & 91.88 & 84.57          & 88.07         & 78.69          & 99.03                   & 64.15 & 57.09          & 60.41         & 43.28          & 87.30                  \\
BIT              & 89.35          & 89.56          & 89.46     & 80.92          & 98.83        & 89.40 & 90.03          & 89.72 & 81.35          & 99.12                  & 56.36  & 62.79          & 59.40         & 42.25          & 85.43                   \\
DASNet          & 90.60          & 91.38           & 90.99            & 83.47          & 99.09          & 88.23                    & 84.62                    & 86.39                  & 76.04                   & 95.30                   & 60.10          &  56.53          &  58.26                  &  41.10                   &  86.25 \\
\hline
SARAS-Net (V1) & 91.48          & 89.35          & 90.40          & 82.49         & 98.95          & 91.41                    &  \textbf{89.58}                    & 90.48                 & 82.62                    & 98.96           & 64.48  & 64.98                     & 64.73         & 47.85           & 88.05                  \\
SARAS-Net (V2) & \textbf{91.97}           & \textbf{91.85} & \textbf{91.91} & \textbf{84.95} & \textbf{99.10}     & \textbf{92.94}   & 89.12           & \textbf{90.99}         & \textbf{83.47}           & \textbf{99.25}         & 67.65  & \textbf{67.51}                    & \textbf{67.58}        & \textbf{51.04}         & \textbf{89.01}          \\ \hline
\end{tabular}}
\label{accuracy}

\end{table*}

\begin{figure}[h]
\centering
\includegraphics[width=0.5\textwidth]{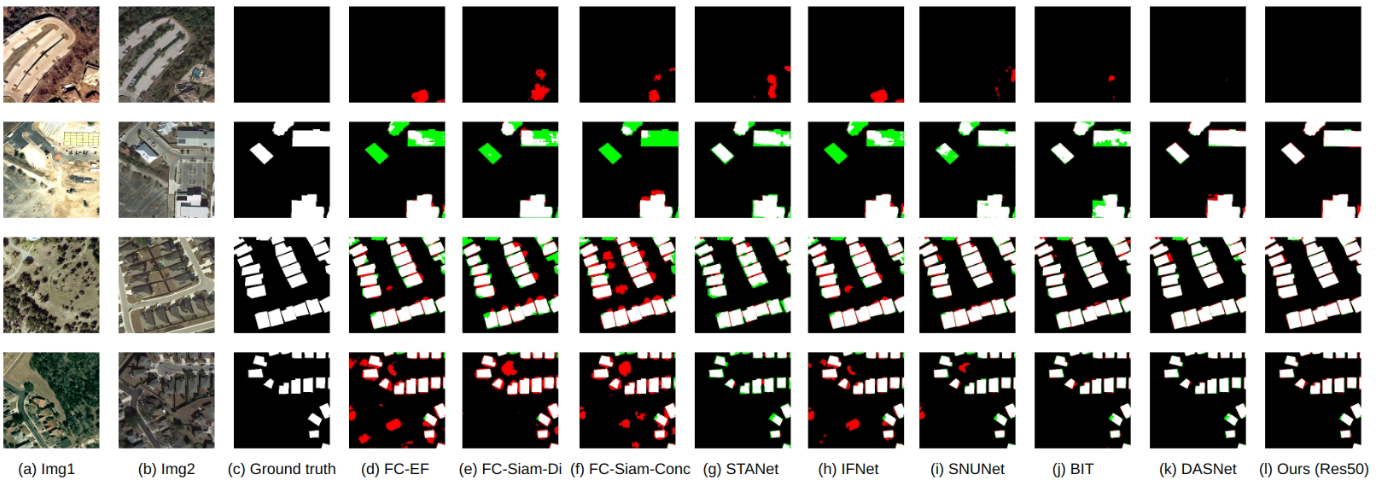}
\footnotesize (a) LEVIR-CD
\includegraphics[width=0.5\textwidth]{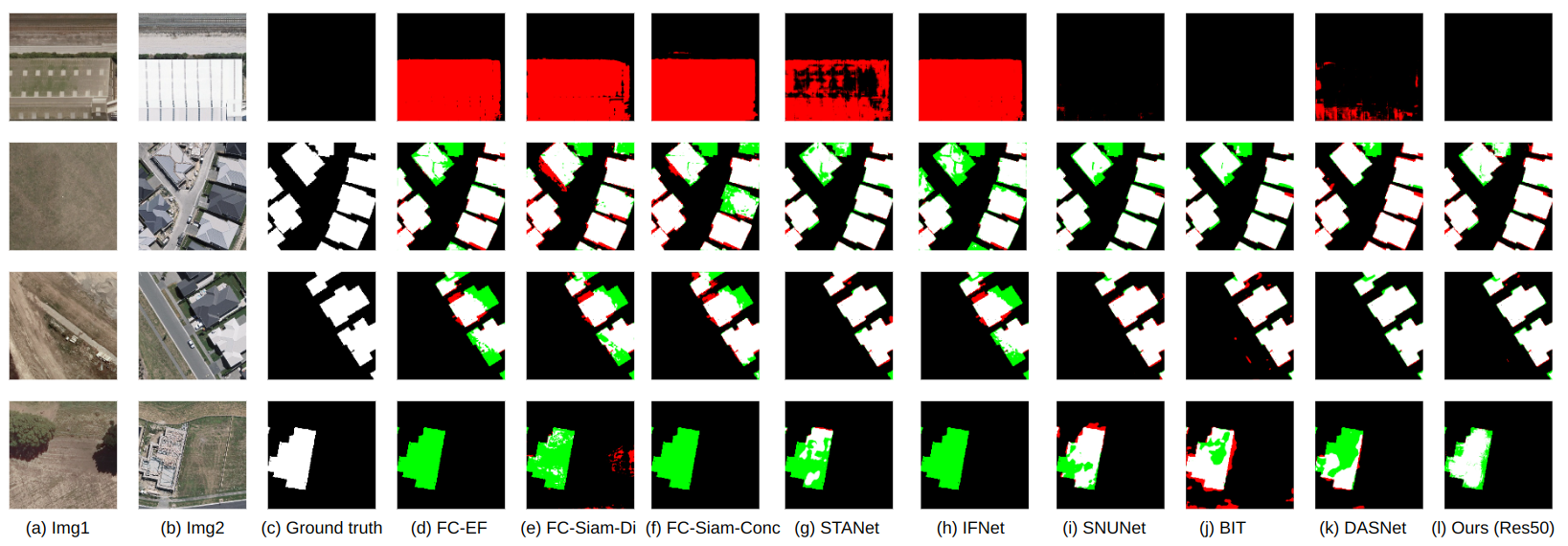}
\footnotesize (b) WHU-CD
\includegraphics[width=0.5\textwidth]{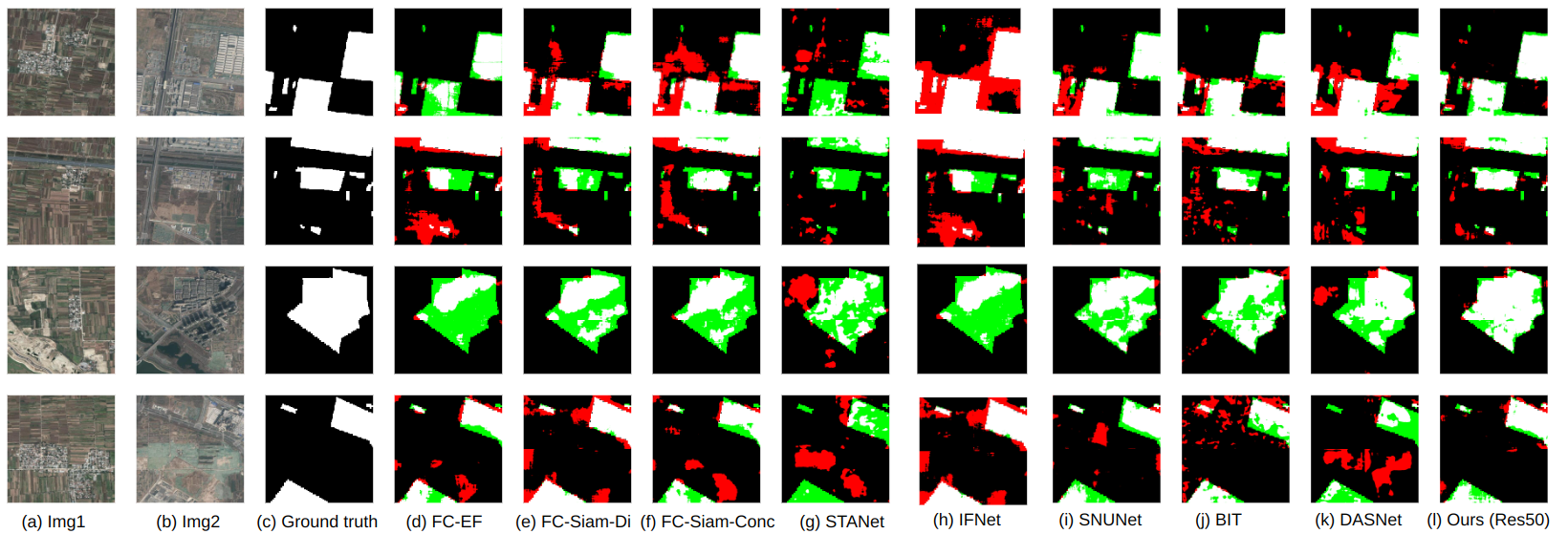}
\footnotesize (c) DSIFN-CD
\caption{Using visualization results to compare our model with other methods on three open datasets; that is, from top to bottom: LEVIR-CD, WHU-CD, and DSIFN-CD.}
\label{dataset_result}
\end{figure}
\vspace{0.2cm}
% \begin{figure}[t]
% \centering
% \includegraphics[height=1.5in]{LEVIR-CD.png}
% \footnotesize (a) LEVIR-CD
% \includegraphics[height=1.5in]{WHU-CD.png}
% \footnotesize (b) WHU-CD
% \includegraphics[height=1.5in]{DSIFN-CD.png}
% \footnotesize (c) DSIFN-CD
% \caption{Using visualization results to compare our model with other methods on three open datasets. From top to bottom: LEVIR-CD, WHU-CD, and DSIFN-CD.}
% \label{dataset_result}
% \end{figure}

\begin{figure}[t]
\centering
\includegraphics[width=0.5\textwidth]{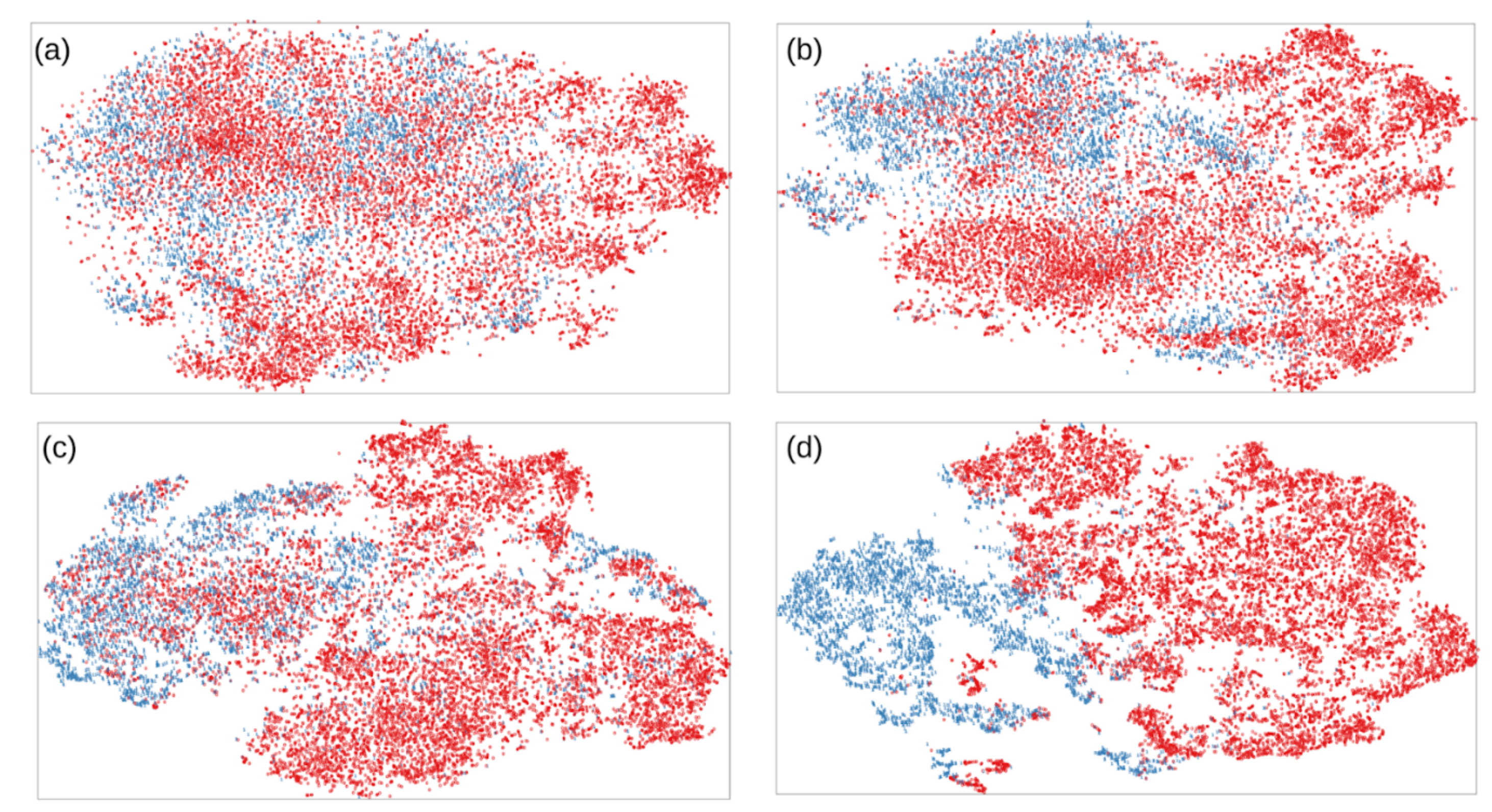}
\caption{ Results of visualization of features enhanced by different modules. (a) Subtraction result without using any module. (b) Result after adding the relation-aware module. (c) Result after adding the scale-attention module. (d) Result after using the cross-transformer module. }
\label{vis_dritub}
\end{figure}
\vspace{0.1cm}

\begin{figure*}[t]
\centering
\includegraphics[width=1\textwidth]{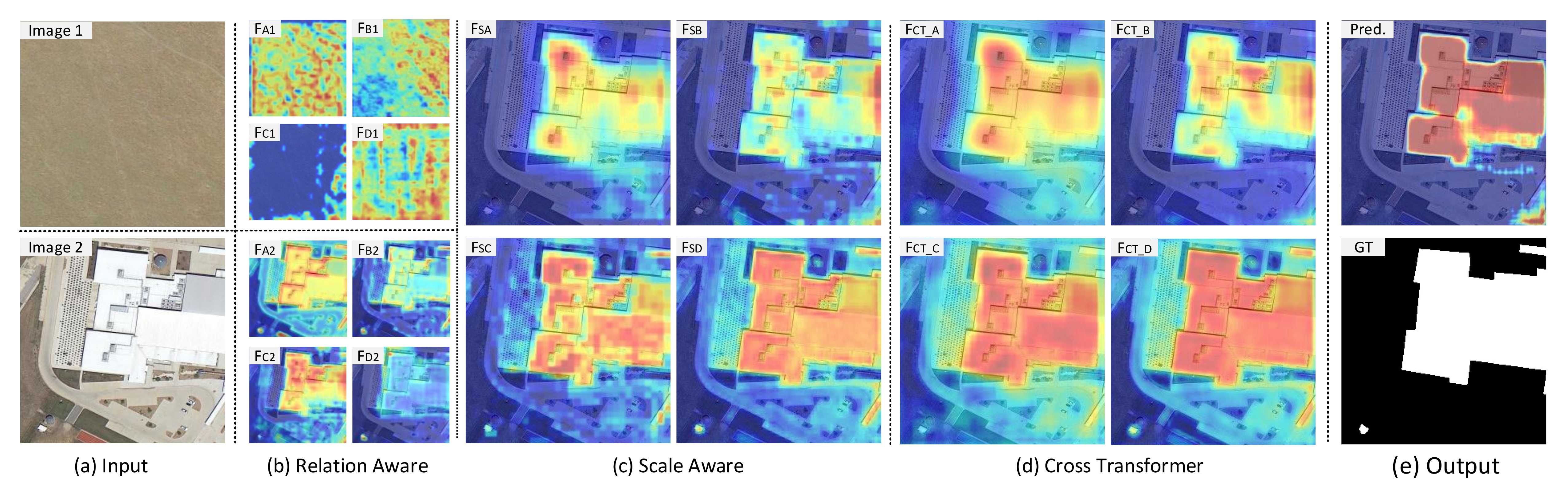}
\caption{ Example of SARAS-Net visualization by Gradcam. Red denotes higher attention values and blue denotes lower values. (a) Two input images. (b) Feature maps generated by the relation-aware module. (c) Subtraction results after adding the scale-aware module. (d) Subtraction results after using the cross-transformer module. (e) Prediction and ground truth.}
\vspace{-0.6cm}
\label{Visualize}
\end{figure*}

\subsection{ Cross transformer module}
\begin{figure}[t]
\centering
\includegraphics[width=0.5\textwidth]{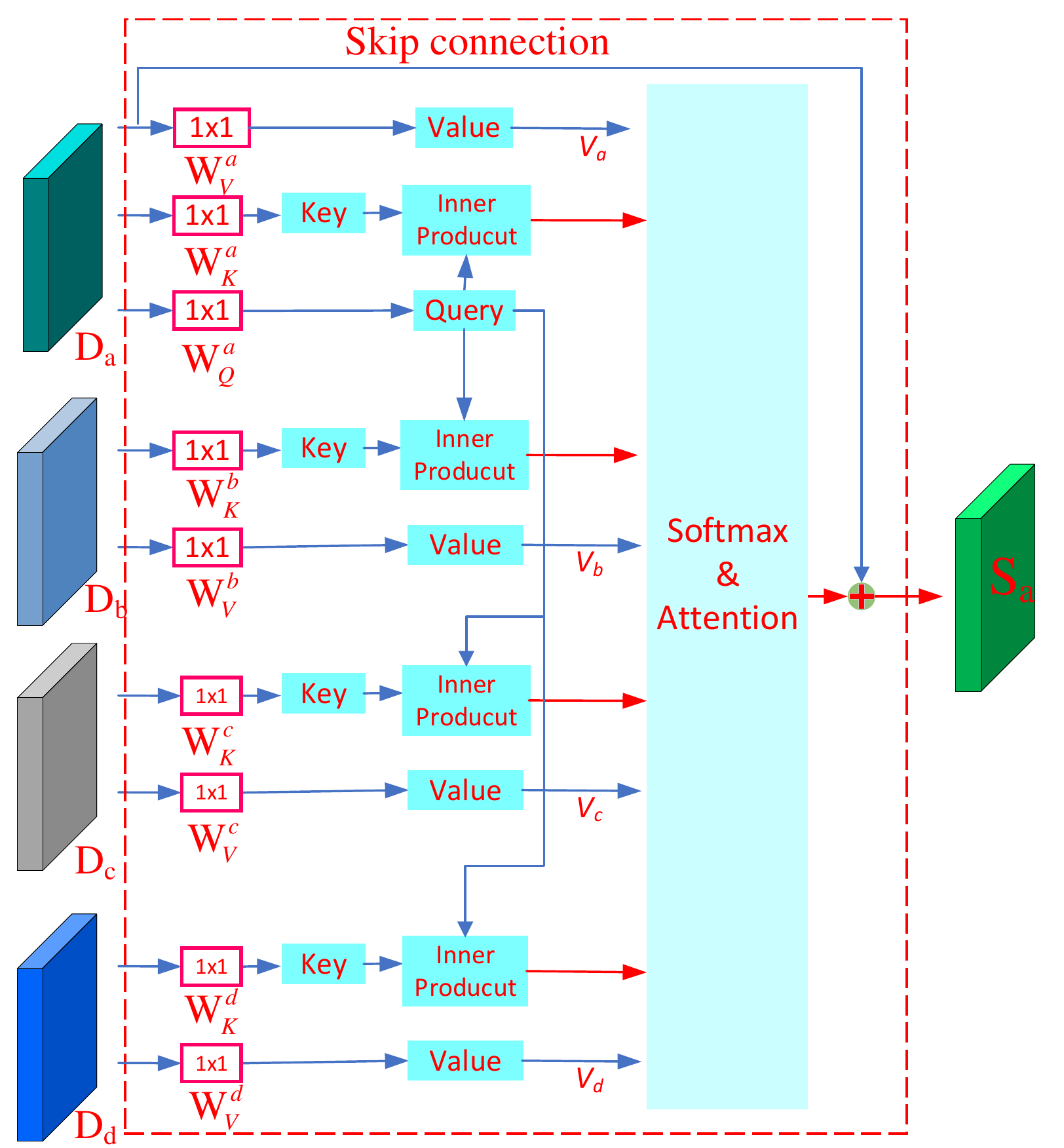}
\caption{ Detailed operations used in CTB (Cross-Transformer Block)}
\label{cross_tran}
\end{figure}
After the scale-aware module enhances each difference feature maps $D_n$, this section proposes a CTB(Cross-Transformer Block), as shown in Figure~\ref{cross_tran}, to generate a scale-aware feature map for better scene detection. Let the inputs of CAB be $D_a$, $D_b$, $D_c$, and $D_d$ which are the subtraction results from different scales. Given  $D_a$, we train three matrices  ${W_Q^a}$, ${W_K^a}$, and ${W_V^a}$ to map  it to the query ${Q_a}$, the key  ${K_a}$, and the value ${V_a}$, respectively.  Similarly, given $D_b$, $D_c$, and  $D_d$, we can train the linear matrices ${W_K^b}$, ${W_V^b}$, ${W_K^c}$, ${W_V^c}$, ${W_K^d}$, and ${W_V^d}$ to get ${K_b}$, ${V_b}$, ${K_c}$, ${V_c}$, ${K_d}$, ${V_d}$, respectively.  Then, based on $Q_a$, we can train the cross-scale attentions $\beta_m$ between it and all keys $V_m$ for $m=a,b,c,d$, where $\beta_m$ is obtained as follows:  

\begin{equation}\label{eq:Attentions} 
\beta^{m} = \frac{Sum({Q_a} \otimes {{K_m} )}}{\sum\limits_{m=a,b,c,d}{Sum({Q_a} \otimes {{K_m} )}}}.
%\vspace{-0.3cm}
\end{equation}

Then, all the $V_m$ will be combined to form $S_a$ as follows:
  \begin{equation}\label{eq:Attentions} 
S^{a} = D_{a} +\sum\limits_{m=a,b,c,d} \beta _m V_m,
\label{SA}
\end{equation}
where a skin connection is used  to avoid the problem of vanish gradient by adding $D_a$ to Eq.(\ref{SA}). For the $n$th layer, by taking $\{D^n_m\}$ as the inputs, CTB will output $S_n$. All $\{S_n\}_{n \neq 1}$ will be normalized to have the same size to $S_1$ with a bilinear interpolation operation. They are further concatenated together and followed by a 3 × 3 convolution to form the change classifier, $G:R^{H\times W\times 4C}\rightarrow R^{H\times W\times 2}$. With $G$, the final predicted change probability map $P$ can be generated via a Softmax function as follow:   

  \begin{equation}\label{eq:Attentions} 
P = Softmax (G(S_1\copyright S_2\copyright S_3\copyright S_4)),
\label{SA}
\end{equation}
where $\copyright$ is a concatenation operation. Algorithm 1 shows the details of our SARAS-Net for change detection. 

\begin{algorithm}
\caption{SARAS-Net for change detection}\label{euclid}
\hspace*{\algorithmicindent} \textbf{Input:}  Two temporal remote sensing images ($X$, $Y$) \\
\hspace*{\algorithmicindent} \textbf{Output:}  Change map $M$
\begin{algorithmic}[1]
\State  \textit{ \textbf{Step1}: Feature Extraction}\\
$F_X$ = $CNN$($X$);   $F_Y$ = $CNN$($Y$);
\State \textit{ \textbf{Step2}: Relation-aware}
\For{layer $n$ }
    \State $ ({\bar F}_X^n, \bar F_Y^n)$ = Relation-Aware-module($F_X^n$,$F_Y^n$); 
    \State $ D_n=abs({\bar F}_X^n- \bar F_Y^n)$;  
\EndFor
\State \textit{\textbf{Step3}: Scale-aware adn Cross-Transformer}
\For{layer $n$ }
{ \\ \  \ \quad ${U}_n$ = Scale-Aware-Attention($D_{n}$); 
    \For{layer $m$ ($m$ $\neq$ $n$) }{ \\
        \qquad \ $D_m^n$ = Channel-wise($U_n$, Resize($D_m$));
    }
    \EndFor
    $S_{n}$ =  CTB($D_1^n$, $D_2^n$, $D_3^n$, $D_4^n$); 
}
\EndFor

\State \textit{ \textbf{Step4}: Change Map Generation}
\State 
\quad $P$ = Softmax($G$ ($S_{1}$\copyright $S_{2}$\copyright $S_{3}$\copyright $S_{4}$));
\State  \Return P 
\end{algorithmic}

\end{algorithm}

\subsection{Network Details}
To extract useful features from two input images, we use two modified backbones, ResNet18 and ResNet50~\cite{He_2016_CVPR}, respectively. Unlike the original ResNet18 and ResNet50, we use four feature maps that are extracted from the last four stages and replace the convolutions used the last two stages with stride 2 to 1 to achieve a speed-accuracy trade-off. In our modified ResNet50, the channel sizes for the four features are 256, 512, 1024, and 2048, respectively. To save GPU computation, we reduce the channels of four feature maps to 64/128/256/512 by 1 $\times$ 1 convolution. We call our model SARAS-Net (V1) and SARAS-Net (V2) when using ResNet18 and ResNet50. 

% \subsection{Loss function}
% We use the cross-entropy loss function to optimize our network parameters. This loss is formulated as follows:
% \begin{equation}
% \ L = -\frac{1}{N}\sum_{n=1}^{N}\left[ (1 - y_{n})\log(1 - {P_{n}} )  \right],
% \end{equation}
% where $y_{n}$ is the class value, which is 0 or 1, representing whether this pixel changes or not, $N$ is the number of pixels, ${P_{n}}$ is the prediction value generated by our network
\subsection{Loss function}
\label{sec:Loss_Function}
We use the cross-entropy loss function to optimize our network parameters and minimize loss value in the training stage. The loss is formulated as follows:
\begin{equation}
L = \frac{1}{N}\sum_{n = 1}^{N} l(P_{n}, Y_{n}),
\end{equation}
where $Y_{n}$ is the class value, which is 0 or 1, representing whether this pixel changes or not, $N$ is the number of pixels, ${P_{n}}$ is the prediction value generated by our network, and $l(P_{n}$, $Y_{n})$ = -$Y_{n}$log($P_{n}$) - $(1 - Y_{n}$)log(1-$P_{n}$) is the cross-entropy loss.
% \begin{figure*}[t]

% \centerline{
% {\footnotesize (a)}
% \includegraphics[width=0.75\textwidth]{LEVIR-CD.png}
% }
% \centerline{
% {\footnotesize (b)}
% \includegraphics[width=0.75\textwidth]{WHU-CD.png}
% }
% \centerline{
% {\footnotesize (c)}
% \includegraphics[width=0.75\textwidth]{DSIFN-CD.png}
% }

% \caption{Using visualization results to compare our model with other methods on three open datasets. From top to bottom: (a) LEVIR-CD, (b) WHU-CD, and (c) DSIFN-CD.}
% \label{dataset_result}
% \end{figure*}

\section{Experiments and Results}

Three datasets, LEVIR-CD~\cite{rs12101662}, DSIFN-CD~\cite{ZHANG2020183}, and WHU-CD~\cite{aaa8444434} were used to evaluate the performance of our model. 
\subsection{Datasets}

\begin{itemize}
\item LEVIR-CD: It contains 637 pairs of remote sensing images with the size 1024 $\times$ 1024 pixels. To reduce the computation and argument training data, the original image was cut into small patches that have 256$\times$256 size. Finally, we obtained 7120/1024/2048 pairs of patches for training/validation/test datasets.
\item WHU-CD: It contains only one pair of aerial images with an image size of 32507 $\times$ 15354 pixels. Then, this image is divided into small non-overlaped patches with the same 256 $\times$ 256 size. In the end, there are 6690/744/744 pairs of patches for the training / validation / test dataset.
\item DSIFN-CD: Its contains changes in roads, buildings, and water. For each pair of images, their sizes are 512$\times$512 pixels and cut into small
non-overlaped patches with 256$\times$ 256 sizes to obtain 14400/1360/192 pairs for training/validation/testing.
\end{itemize}

\subsection{Implementation Details}

In the training stage, we use three NVIDIA Tesla V100 GPU to implement our model and stochastic gradient descent (SGD) to optimize our model parameters. We set the momentum to 0.9 and the weight decay to 0.0005. Initially, the learning rate is set to 0.05 and decays 0.1 times for every 50 epochs. For each epoch, we use data augmentation to obtain higher accuracy, including rescale, crop, flip, and Gaussian blur during training and use the validation dataset to choose the best training weight. Finally, we evaluate our model accuracy on the test dataset.

\subsection{Experiments on dataset}
We compared our model with eight SoTA methods, including FC-EF~\cite{aa8451652}, FC-Siam-Di, FC-Siam-Conc, SNUNet, STANet~\cite{rs12101662}, IFNet~\cite{ZHANG2020183} , ISNet~\cite{9772654}, BIT, and DASNet~\cite{DASNet2021}.

 Table~\ref{accuracy} illustrates performance comparisons with the SoTA methods in the three data sets, respectively. The metrics contain precision, recall, F1 score, intersection over union (IoU) of the change category, and general accuracy.  Clearly, our model outperforms all SoTA methods. To visualize the prediction results, the results of different methods on the above three datasets are shown in Figure~\ref{dataset_result}. Here, the white color is for true positive, the black color is for true negative, the red color is for false positive, and the green color is for false negative. 
 
 To evaluate the performance of each module, we projected high-dimensional features onto 2D maps with two colors to represent the class of pixels (see Figure~\ref{vis_dritub}). First, the subtraction result between two input images, shown in Figure~\ref{vis_dritub}(a), illustrates the changed and unchanged pixels are mixed together. Second, after the self-attention and cross-attention modules, points with the same class become closer (see Figure~\ref{vis_dritub}(b)). Third,
as shown in Figure~\ref{vis_dritub}(c), the scale-attention module makes points with different classes become farther. Finally, as shown in Figure~\ref{vis_dritub}(d), the
cross-transformer block fuses features from different layers to make changed and unchanged pixels more easily separated.

In order to better understand our model, we use gradcam~\cite{jacobgilpytorchcam} to visualize each module. Figure~\ref{Visualize}(a) shows the two input images. Figure~\ref{Visualize}(b) shows the feature map after adding the relation-aware module. Figure~\ref{Visualize}(c) shows the subtraction results after using the scale-aware module. From Figure~\ref{Visualize}(d), we observe that the noise in features is alleviated through the cross-transformer module. The final prediction map is shown in Figure~\ref{Visualize}(e).

\subsection{Ablation study}
\begin{figure}[h]
\centering{
\includegraphics[width=0.5\textwidth]{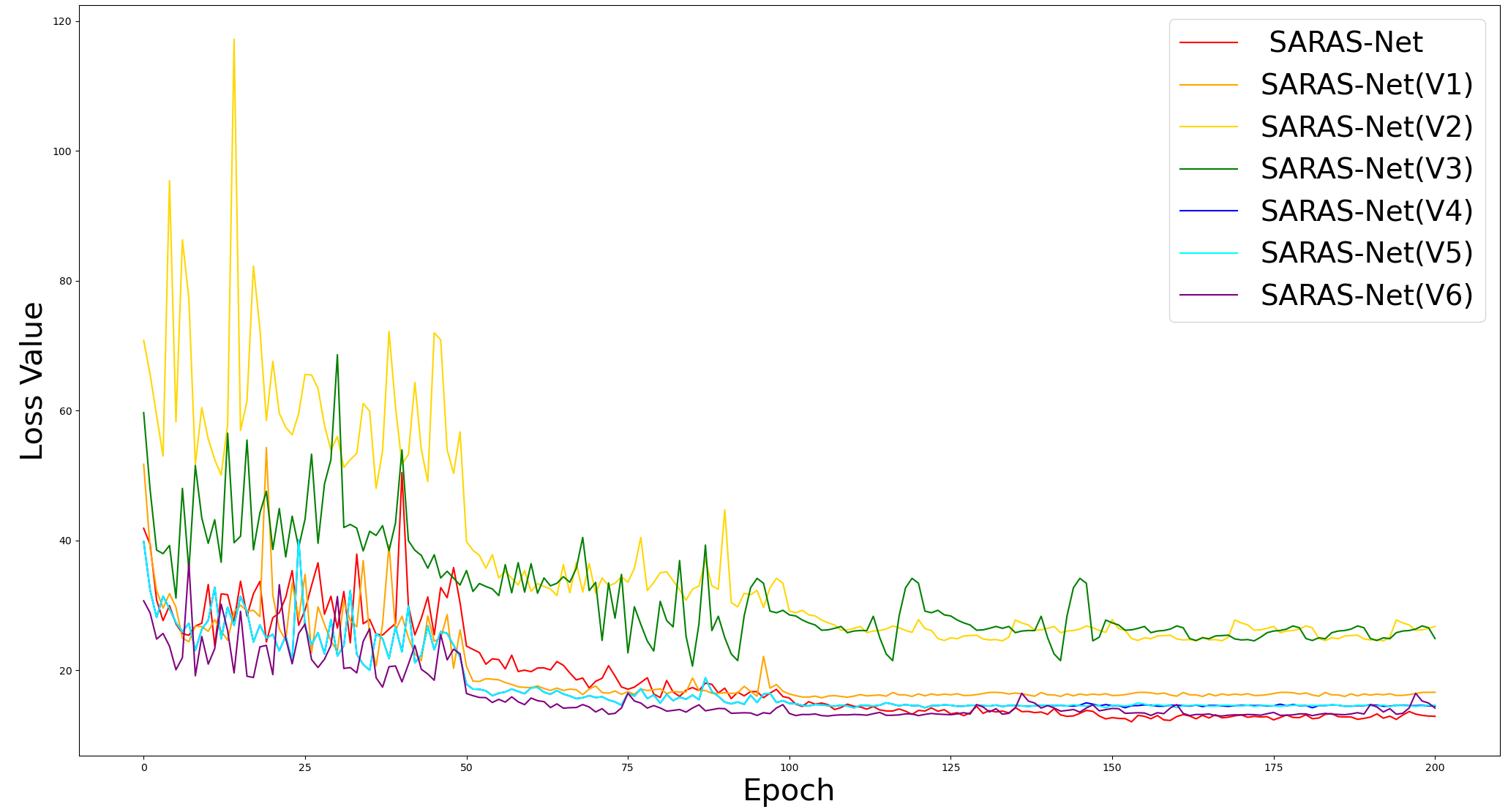}
}
\caption{ Ablation study of training loss. Each line represents an ablation study from Table~\ref{ablation_study}.}
\label{training_loss}
\end{figure}

\textbf{SARAS-Net analysis}. We performed an ablation study for our model on LEVIR-CD data set and changed different modules in sequence to evaluate the performance of each module. First, from Table~\ref{ablation_study}, we can conclude that when we remove the relation-aware (RA) module, the number of parameters is reduced by nearly half and the FLOPs also decrease. However, the performance after removing the RA module is worse than after removing other modules. Thus, if we want a light network, we will remove the RA module. Second, we observe that the cross transformer module has higher performance since it can perfectly fuse different scale features.  Figure~\ref{training_loss} shows the changes in the loss value at each epoch during training.  We observe that our model without removing any module is easier to converge.

% \begin{table}[]
% \caption{ Ablation study of our model efficiency and effectiveness, including using the number of parameters (Params.), floating-point operations per second (FLOPs), F1-score, and intersection over union (IoU). RA is relation-aware module. SA is scale-aware module. CT is cross transformer module.}
% \scalebox{0.7}{\begin{tabular}{lccccccc}
% \hline
% Model     & RA & SA & CT & Param.(M) & FLOPs.(G) & F1    & IoU   \\ \hline
% SARAS-Net(v1) & \ding{53}  & \ding{53}  & $\surd$   & 32.92     & 73.88     &  90.63     & 82.86      \\
% SARAS-Net(v2) & \ding{53}  & $\surd$    & \ding{53}  & 49.55     & 145.73    &  90.49     & 82.63      \\
% SARAS-Net(v3) & $\surd$   & \ding{53}  & \ding{53}  & 100.39    & 127.15    &  90.48     & 82.62      \\
% SARAS-Net(v4) & \ding{53}  & $\surd$    & $\surd$   & 54.09     & 164.96    & 91.45 & 84.26 \\
% SARAS-Net(v5) & $\surd$   & \ding{53}  & $\surd$    & 117.70    & 226.85    & 91.11 & 83.68  \\
% SARAS-Net(v6) & $\surd$   & $\surd$   & \ding{53}  & 117.83    & 212.23    & 90.92 & 83.36  \\
% SARAS-Net & $\surd$   & $\surd$   & $\surd$    & 122.36    & 231.46    & \textbf{91.79} & \textbf{84.82} \\ \hline
% \end{tabular}}
% \label{ablation_study}
% \end{table}

\vspace{0.1cm}
\begin{table}[h]
\caption{ Ablation study of the effects when different modules are added to our model, where RA is the relation-aware module, SA is the scale-aware module, and CT is the cross-transformer module.}
\scalebox{0.7}{\begin{tabular}{lccccccc}
\hline
Model     & RA & SA & CT & Param.(M) & FLOPs.(G) & F1    & IoU   \\ \hline
SARAS-Net(v1) & \ding{53}  & \ding{53}  & $\surd$   & 32.33     & 60.64     &  90.63     & 82.86      \\
SARAS-Net(v2) & \ding{53}  & $\surd$    & \ding{53}  & 37.02     & 92.57    &  90.49     & 82.63      \\
SARAS-Net(v3) & $\surd$   & \ding{53}  & \ding{53}  & 47.46    & 76.91    &  90.48     & 82.62      \\
SARAS-Net(v4) & \ding{53}  & $\surd$    & $\surd$   & 42.94     & 111.77    & 91.26 & 83.39 \\
SARAS-Net(v5) & $\surd$   & \ding{53}  & $\surd$    & 52.36    & 108.83    & 91.11 & 83.68  \\
SARAS-Net(v6) & $\surd$   & $\surd$   & \ding{53}  & 53.16    & 128.03    & 90.92 & 83.36  \\
SARAS-Net & $\surd$   & $\surd$   & $\surd$    & 56.89    & 139.9    & \textbf{91.91} & \textbf{84.95} \\ \hline
\end{tabular}}
\label{ablation_study}
\end{table}

\begin{figure}[h]
\centering
\includegraphics[width=0.5\textwidth]{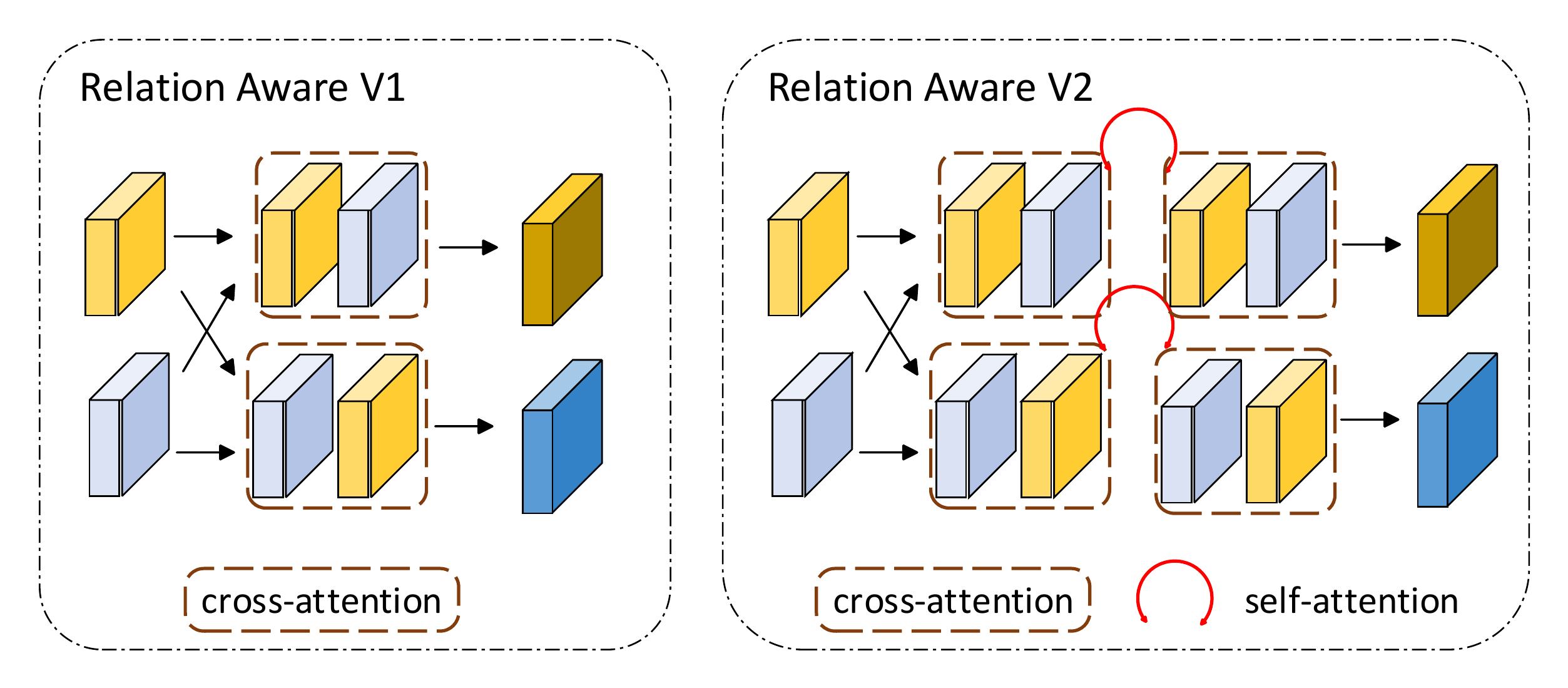}
\caption{ Ablation study of relation-aware module. Relation-aware V1 only uses cross-attention. Relation-aware V2 uses cross-attention and self-attention.}
\label{Relation_study}
\end{figure}

\begin{figure}[t]
\centering
\includegraphics[width=0.5\textwidth]{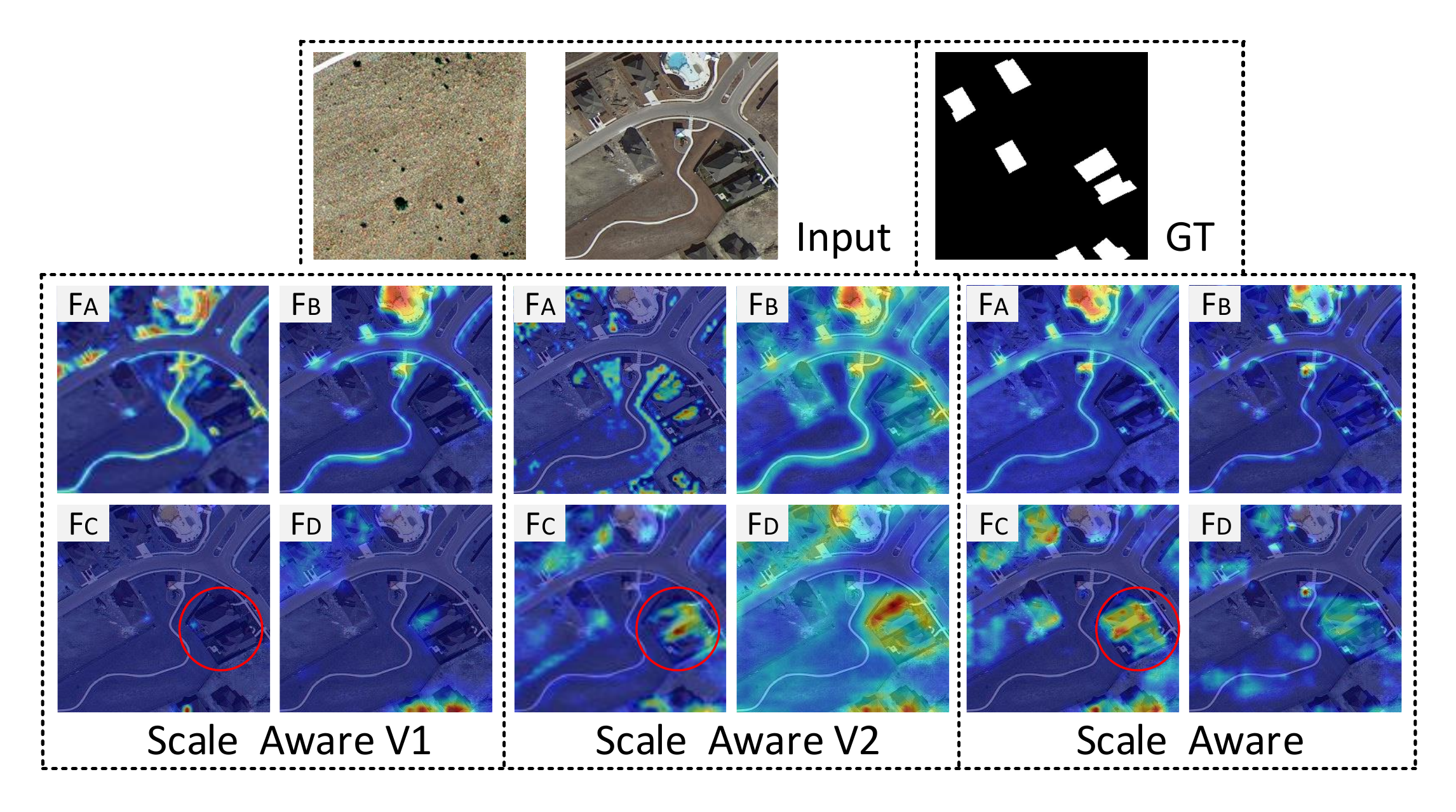}
\caption{ Ablation study of the relation-aware module. V1 uses only cross-attention and V2 uses both cross-attention and self-attention.}
\vspace{0cm}
\label{Relation_study_image}
\end{figure}
\vspace{0.2cm}

\begin{table}[h]
\caption{ Ablation study of the relation-aware module, where CA denotes cross-attention, CSA the cross-self-attention, and SA the self-attention.}
\scalebox{0.7}{\begin{tabular}{lccccccc}
\hline
Model     & CA & CSA & SA & Param.(M) & FLOPs.(G) & F1    & IoU   \\ \hline
SARAS-Net & $\surd$   & \ding{53}   & \ding{53}  & 49.22     & 131.77    & 91.54      & 84.41      \\
SARAS-Net & $\surd$   & \ding{53}  & $\surd$   & 56.89    & 139.90    &  90.56 & 82.76      \\
SARAS-Net & $\surd$   & $\surd$    & \ding{53}  & 56.89    & 139.90    & \textbf{91.91} & \textbf{84.95} \\ \hline
\end{tabular}}
\label{ablation_relation_aware}
\end{table}
\vspace{0.2cm}

\textbf{Relation-aware module analysis.} To evaluate the ablation study of the relation-aware module, we performed different attention mechanisms. As shown in Figure~\ref{Relation_study}, the first relation-aware version only uses cross-attention, and the second uses cross-attention initially and then replaces cross-self-attention with self-attention. From Table~\ref{ablation_relation_aware}, we can observe that the first version is lighter but has a worse performance. To better understand this ablation study, we visualize the different scale features using the relation-aware module by Gradcam in Figure~\ref{Relation_study_image}. Compared to other ablation studies of the relation-aware module, we observe that the original relation-aware module pays more attention to regions with changes. For example, as shown in Figure~\ref{Relation_study_image}, the red circle in the original module feature $F_C$ performs better.  

\vspace{0.1cm}
\section{Conclusion}

This paper proposed a scale- and relation-aware siamese network for change detection to achieve SoTA accuracy on the LEVIR-CD, WHU-CD, and DSIFN-CD datasets. More accurately, our model obtains significant improvements in F1 scores in these datasets, respectively, 2.45, 1.27, and 4.63 points. Our method can solve the key problems of change detection encountered with most existing methods. For example, the relation-aware and scale-aware modules can resolve boundary noise generated by objects of different scales and enhance the features of interactive information.  In addition, we fuse the different scale features using the cross-transformer module to get a better representation for change detection. Except for these, our main contribution is to propose a new model, which performs operations before and after feature subtraction. Through experimental evidences, our model structure has been proven to be useful in this issue.               
\vspace{0.1cm}

% \begin{figure*}[htbp]
% \centering
% \includegraphics[width=1\textwidth]{LEVIR-CD.png}
% \footnotesize (a) LEVIR-CD
% \includegraphics[width=1\textwidth]{WHU-CD.png}
% \footnotesize (b) WHU-CD
% \includegraphics[width=1\textwidth]{DSIFN-CD.png}
% \footnotesize (c) DSIFN-CD
% \caption{Using visualization results to compare our model with other methods on three dataset. From top to bottom: LEVIR-CD, WHU-CD, and DSIFN-CD. White color for true positive, black color for true negative, red color for false positive, and green color for false negative.}
% \label{dataset_result}
% \end{figure*}

\section{Acknowledgments}
This work was partly supported by National Science and Technology Council, Taiwan (Grant Number: MOST 109-2221-E-009-116-MY3, 110-2221-E-A49-132-MY3 and 110-2634-F-A49-006).

\bibliographystyle{aaai23}
\bibliography{aaai23}

\end{document}